\documentclass[runningheads]{llncs}
\usepackage[T1]{fontenc}
\usepackage{graphicx}
\usepackage{diagbox,multirow,epsfig,fbox,amsfonts,amsmath,multicol,enumitem,algorithm,pifont}
\usepackage{arydshln}
\usepackage{amssymb}
\usepackage{booktabs}
\usepackage{bm}
\usepackage[caption=false,font=footnotesize]{subfig}
\usepackage{array}

\usepackage{algpseudocode}
\usepackage[subfigure,titles]{tocloft}
\begin{document}
\title{Value-Guided Iterative Refinement and the DIQ-H Benchmark for Evaluating VLM Robustness}
%
%

\author{
Hanwen Wan\inst{1,3} \and
Zexin Lin\inst{1,3} \and
Yixuan Deng\inst{1,3} \and
Xiaoqiang Ji\inst{1,2,3,\dag}}

\authorrunning{H. Wan et al.}

\institute{The School of Science and Engineering, The Chinese University of Hong Kong, Shenzhen, China. \and
The School of Artificial Intelligence, The Chinese University of Hong Kong, Shenzhen, China. \and
The Shenzhen Institute of Artificial Intelligence and Robotics for Society, Shenzhen, China. \\
\textsuperscript{$\dagger$}Corresponding author: jixiaoqiang@cuhk.edu.cn }

\maketitle

\let\thefootnote\relax 
\footnotetext{This work was partially supported by National Natural Science Foundation of China (Grant No. 62441619), Guangdong Basic and Applied Basic Research Foundation (Grant No. 2023A1515012883, 2022A1515110411, and 2024A1515240009), Shenzhen Science and Technology Program (Grant No. JCYJ20240813113604006, and KJZD20240903095730039), Science and Technology Planning Project of Guangxi Province under Grant (AA23062031-2, AA23062073-2)}

\begin{abstract}
Vision-Language Models (VLMs) are essential for embodied AI and safety-critical applications, such as robotics and autonomous systems. However, existing benchmarks primarily focus on static or curated visual inputs, neglecting the challenges posed by adversarial conditions, value misalignment, and error propagation in continuous deployment. Current benchmarks either overlook the impact of real-world perturbations, or fail to account for the cumulative effect of inconsistent reasoning over time. 
To address these gaps, we introduce the Degraded Image Quality Leading to Hallucinations (DIQ-H) benchmark, the first to evaluate VLMs under adversarial visual conditions in continuous sequences. DIQ-H simulates real-world stressors including motion blur, sensor noise, and compression artifacts, and measures how these corruptions lead to persistent errors and misaligned outputs across time. The benchmark explicitly models error propagation and its long-term value consistency.
To enhance scalability and reduce costs for safety-critical evaluation, we propose the Value-Guided Iterative Refinement (VIR) framework, which automates the generation of high-quality, ethically aligned ground truth annotations. VGIR leverages lightweight VLMs to detect and refine value misalignment, improving accuracy from 72.2\% to 83.3\%, representing a 15.3\% relative improvement. The DIQ-H benchmark and VGIR framework provide a robust platform for embodied AI safety assessment, revealing vulnerabilities in error recovery, ethical consistency, and temporal value alignment.

\end{abstract}

\section{Introduction}

Vision-Language Models (VLMs) are increasingly deployed in safety-critical applications such as autonomous driving, robotic manipulation~\cite{wan2025toward}, and healthcare assistance, where they must interpret continuous visual streams under imperfect conditions while maintaining reliable, ethically sound, and value-grounded reasoning. A fundamental challenge is hallucination~\cite{zhang2024toolbehonest}—the tendency to fabricate non-existent objects or attributes—which becomes particularly dangerous when errors compound over time in sequential reasoning tasks. In safety-critical contexts, transient visual degradation can trigger persistent erroneous beliefs that contaminate downstream decisions, potentially leading to systematic failures in value-sensitive decision contexts where ethical consistency and trustworthy operation are paramount.

Current evaluation paradigms suffer from three critical gaps: (1) Static focus: Benchmarks like LLaVA-Bench~\cite{liuVisualInstructionTuning2023a} and MME~\cite{fuMMEComprehensiveEvaluation2024} assess only single-frame understanding; (2) Temporal blindness: Hallucination benchmarks (POPE~\cite{liEvaluatingObjectHallucination2023a}, ToolBeHonest~\cite{zhang2024toolbehonest}) evaluate isolated responses without modeling error propagation or ethical consistency over time; (3) Idealized inputs: Video benchmarks~\cite{zhang2024unveilingtapestryconsistencylarge} assume pristine quality, ignoring real-world degradation from motion blur, sensor noise, and compression artifacts. These limitations obscure a critical failure mode: cognitive inertia, where hallucinations induced by transient degradation persist even after visual quality recovers, compromising reliable and ethically consistent operation.

We introduce DIQ-H (Degraded Image Quality leading to Hallucinations), the first benchmark to evaluate VLMs under dynamic visual degradation in temporal sequences, with explicit attention to preserving sound reasoning under challenging conditions. DIQ-H addresses the above gaps through: (1) Physics-based degradation simulation applying realistic motion blur, Poisson-Gaussian noise, and H.265 compression; (2) Temporal task design with multi-turn Q\&A probing error propagation, recovery, and consistent value-grounded reasoning; (3) Adaptive difficulty calibration that stress-tests model limits in reliability-critical scenarios. To enable scalable annotation for safety-critical evaluation, we propose VIR (Value-guided Iterative Refinement), which generates reliable pseudo-ground-truth using lightweight VLMs with value-sensitive uncertainty filtering, achieving 15.3\% accuracy improvement over direct annotation. Our main contributions are:
\begin{itemize}
\item \textbf{DIQ-H Benchmark}: Systematic evaluation of VLM robustness to sequential degradation and hallucination propagation in dynamic video environments, with explicit measurement of consistent, value-grounded reasoning and ethical reliability under adversarial visual conditions.
\item \textbf{Multi-Agent Generation Framework}: Scalable pipeline combining physics-based degradation simulation, temporal task design probing error recovery, and adaptive difficulty calibration for stress-testing model limits in safety-critical and value-sensitive scenarios.
\item \textbf{VIR Annotation Framework}: Cost-effective method for high-quality, ethically aligned ground-truth synthesis via value-sensitive iterative refinement and uncertainty filtering, reducing reliance on human reviewers or GPT-4o supervision while ensuring reliable annotation quality.
\end{itemize}

\section{Related Work}

With the increasing use of VLM, challenges regarding their stability, accuracy, and controllability~\cite{wantowards} have become more prominent. Benchmarking VLM has therefore become a focal point of multimodal intelligence research.

Early benchmarks primarily assessed semantic comprehension over static images. Datasets such as VQA v2~\cite{goyalMakingVQAMatter2017} and the RefCOCO series~\cite{maoGenerationComprehensionUnambiguous2016} focused on tasks like visual question answering and grounding. TextVQA~\cite{singhVQAModelsThat2019} further introduced textual content within images to probe models' OCR capabilities. VizWiz~\cite{gurariVizWizGrandChallenge2018} extended this direction by simulating assistive scenarios involving visually impaired users. In addition, NLVR2~\cite{suhrCorpusReasoningNatural2019} examines models making logical judgments between two images, and ScienceQA~\cite{luLearnExplainMultimodal2022}, MMMU~\cite{yueMMMUMassiveMultidiscipline2024}, and others further extend to scientific reasoning and integrated multidisciplinary assessment. More recent efforts~\cite{liuVisualInstructionTuning2023a,fuMMEComprehensiveEvaluation2024} consolidated these tasks into unified platforms, serving as representative benchmarks for evaluating instruction following, object recognition, and semantic alignment on clean, static images.

In parallel, hallucination detection has become an increasingly important subfield~\cite{wan2025embodiedagent}. This has led to the creation of dedicated benchmarks designed to examine the alignment between model output and visual input. Discriminative benchmarks typically formulate the task as binary classification that assesses whether a model correctly describes an object or attribute present in the image. Notable benchmarks in this category include POPE~\cite{liEvaluatingObjectHallucination2023a}, which uses yes or no questions on object presence, NOPE~\cite{loveniaNegativeObjectPresence2024}, which further increases scale and granularity. BEAF~\cite{ye-binBEAFObservingBEforeAFter2024}, which introduces before-after scene edits to test sensitivity to changes, further probe robustness under structured perturbations. Furthermore, structured annotation protocols proposed in THRONE~\cite{kaulTHRONEObjectbasedHallucination2025} and HQH~\cite{yanEvaluatingQualityHallucination2024} aim to enhance the reproducibility and transparency of hallucination evaluations.

Generative benchmarks, in contrast, allow models to produce open-ended responses that are later scored based on correctness, relevance, or consistency.  Benchmarks such as GAVIE~\cite{liuMitigatingHallucinationLarge2024}, HaELM~\cite{wangEvaluationAnalysisHallucination2023}, and Bingo~\cite{cuiHolisticAnalysisHallucination2023} assess object-level correctness while also identifying systematic biases and sensitivity to prompt phrasing or visual cues. M-HalDetect~\cite{gunjalDetectingPreventingHallucinations2024} introduces reward model scoring to detect hallucination through indirect feedback, while MMHal-Bench~\cite{sunAligningLargeMultimodal2023} incorporates alignment strategies from language modeling into the multimodal domain. Notably, HalEval~\cite{jiangHalEvalUniversalFinegrained2024} integrate both discriminative and generative frameworks to offer a comprehensive assessment.

Although existing benchmarks have advanced VLM evaluation, they remain limited by static inputs, idealized visual conditions, and isolated testing that overlooks temporal error propagation. To address these gaps, we introduce DIQ-H, a benchmark that evaluates hallucination behaviors under continuous visual degradation in real-world video sequences. By interleaving systematic distortions with multi-turn question-answering, DIQ-H assesses whether hallucinations persist, amplify, or self-correct over time—providing a critical foundation for evaluating safety and reliability in realistic, long-term deployments.

\section{Methodology}\label{sec:method}



\begin{figure*}[!t]
\centering
\includegraphics[width=\textwidth]{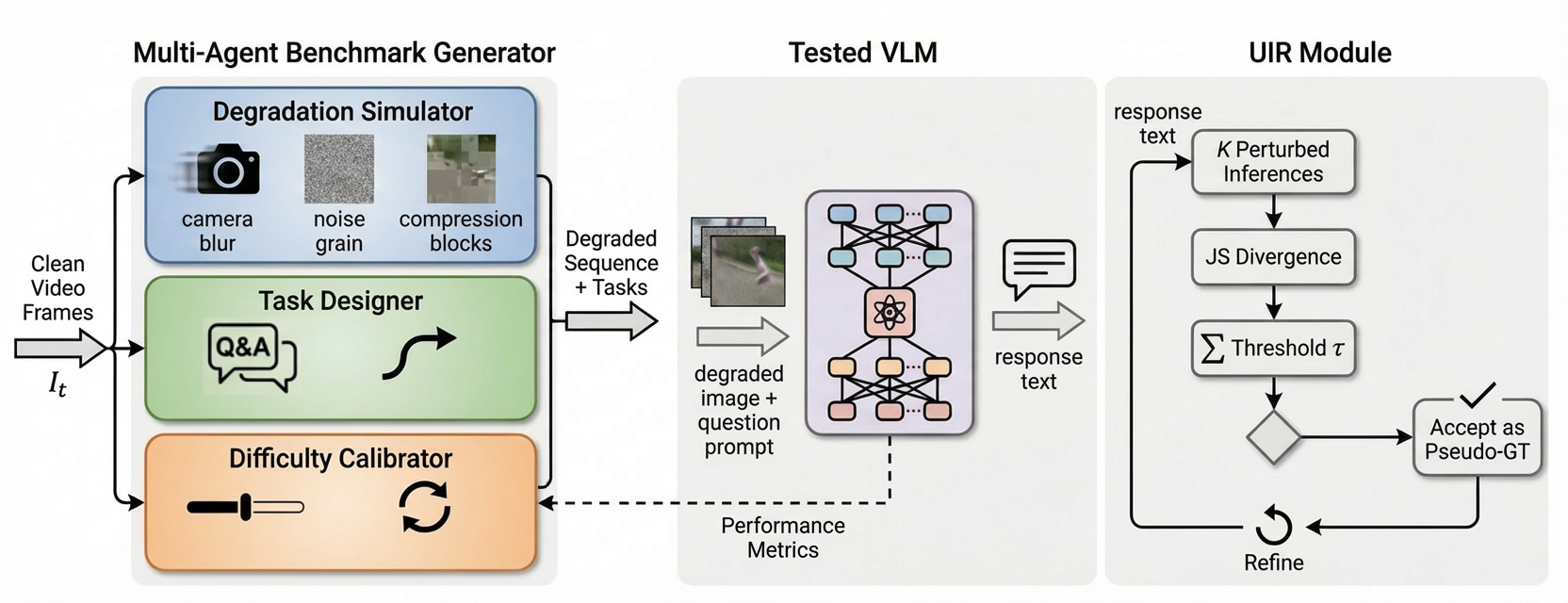}
\caption{Overview of the DIQ-H evaluation framework. The Multi-Agent Benchmark Generator (left) creates temporally degraded sequences through coordinated Degradation Simulator, Task Designer, and Difficulty Calibrator agents. The Tested VLM (center) processes these sequences, with performance metrics fed back for adaptive difficulty control. The VIR Module (right) generates reliable pseudo-ground truth annotations through uncertainty-guided filtering.}
\label{fig:framework}
\end{figure*}

\subsection{Multi-Agent Benchmark Generation}

\begin{figure*}[!t]
\centering
\includegraphics[width=\textwidth]{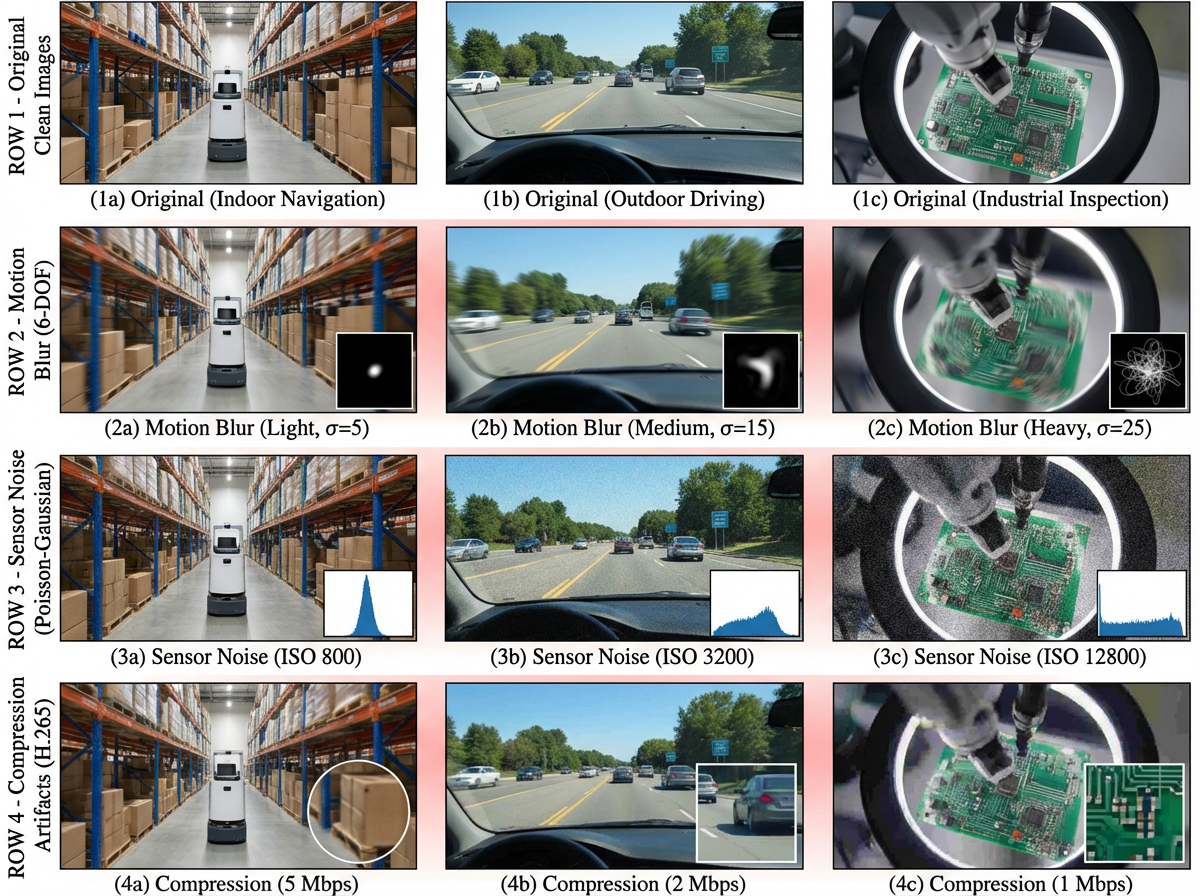}
\caption{Visualization of the three primary degradation types at varying severity levels. Each column shows the same scene under increasing degradation intensity. Motion blur (Row 2) introduces directional streaking from simulated camera motion. Sensor noise (Row 3) adds ISO-dependent Poisson-Gaussian artifacts. Compression (Row 4) produces blocking and ringing from aggressive H.265 encoding.}
\label{fig:degradation}
\end{figure*}

To evaluate VLM robustness under temporally degraded conditions, we introduce a multi-agent framework coordinating three specialized agents: the Degradation Simulator, Task Designer, and Difficulty Calibrator. These agents jointly produce progressive video sequences with controlled corruption levels and adaptive task complexity, reflecting operational challenges in safety-critical robotic and autonomous systems.

The benchmark employs a temporal interaction structure where each test case is framed as multi-turn dialogue over a sequential video stream. Initial frames are degraded via physics-based corruption—motion blur, Poisson-Gaussian noise, and H.265 compression—requiring the model to reason under impaired conditions. As the sequence progresses, clean frames are introduced, but the goal is not to test processing of clear inputs, but to assess recovery from persistent hallucinations induced by earlier degradation. We formalize this as latent error propagation: given input sequence $\{(I_t, Q_t)\}_{t=1}^T$, responses for $t > k$ (clean frames) still reflect errors introduced during $t \leq k$ (degraded frames), even without ongoing visual corruption. This mirrors robotics scenarios where transient visual failures cause incorrect inferences that persist through future actions.

The degradation simulator models real-world quality deterioration through physics-informed transformations~\ref{fig:degradation}. Motion blur uses a 6-DOF point spread function $\mathcal{K}_{\text{motion}}(\boldsymbol{\theta}_t) * I_t + \varepsilon$ with camera motion parameters $\boldsymbol{\theta}_t \in \mathbb{R}^6$. Sensor noise applies an ISO-dependent Poisson-Gaussian model $\mathcal{P}(g \cdot I_t) + \mathcal{N}(0, \sigma^2)$. Compression artifacts are introduced via H.265/HEVC encoding $\text{HEVC}(I_t; B_t)$ at bitrate levels $B_t \in \{1,\dots,5\}$ Mbps.

These three degradation families are chosen to emulate the primary noise sources encountered in embodied AI perception pipelines. Specifically, (1) Motion blur directly corresponds to rapid ego-motion or object dynamics in autonomous driving and drone navigation, which often renders frame-by-frame object detection brittle. (2) Sensor noise (Poisson-Gaussian) replicates the low-light performance limits of CMOS sensors during nighttime robotics operation or in subterranean environments. (3) Compression artifacts reflect bandwidth-constrained streaming scenarios, such as remote teleoperation over 4G/5G networks or long-term onboard data logging where aggressive encoding is employed. By focusing on these foundational distortion axes, DIQ-H probes the model's resilience to the most statistically prevalent and systemically impactful visual perturbations in continuous deployment.

The task designer dynamically generates temporally coherent queries conditioned on temporal memory and object trajectories. Rather than isolated tasks, the agent constructs fragile reasoning contexts—multi-modal chains probing spatial changes, temporal descriptor consistency, and re-identification errors under degradation. Prompts combine template-based generation with paraphrasing for linguistic diversity while maintaining logical consistency with visual evidence.

The difficulty calibrator implements closed-loop adaptive control via degradation strength $\lambda_t = \alpha \cdot \text{EPI}_{t-1} + \beta \cdot \text{HR}_{t-1}$, mapping to motion blur, ISO level, and bitrate. This creates a stress-testing environment that automatically scales difficulty with model performance, exposing weaknesses in uncertainty handling, error recovery, and consistent semantic grounding across evolving streams.

\subsection{Value-Guided Iterative Refinement (VIR)}

\begin{figure*}[!t]
\centering
\includegraphics[width=\textwidth]{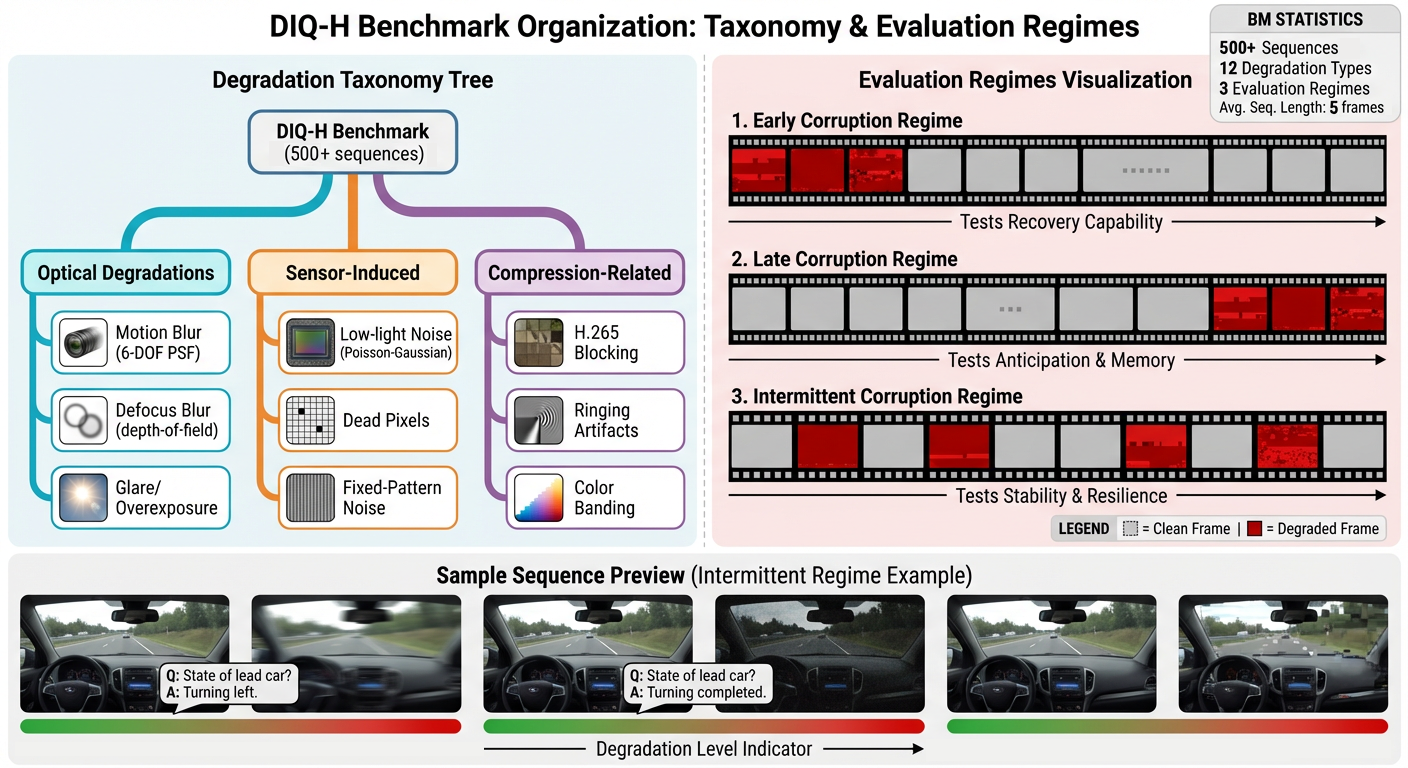}
\caption{Structure of the DIQ-H benchmark. Left: Hierarchical taxonomy of 12 degradation types organized into optical, sensor-induced, and compression-related categories. Right: Three evaluation regimes—early corruption (testing recovery), late corruption (testing memory retention), and intermittent corruption (testing stability)—each designed to probe different aspects of temporal robustness.}
\label{fig:benchmark}
\end{figure*}

To reduce reliance on costly GPT-4o or human annotation for safety-critical evaluation, we propose Value-Guided Iterative Refinement (VIR). The framework generates ethically reliable pseudo-ground truth using lightweight VLMs by identifying outputs where model beliefs are unstable or inconsistent with value-grounded reasoning—signals captured through uncertainty quantification.

Uncertainty is introduced via two perturbation strategies: (1) controlled noise with Gaussian blur injection into input images and (2) stochastic dropout during decoding. Given $K$ perturbed forward passes yielding distributions $\{p^{(i)}(\mathbf{h})\}_{i=1}^K$, we compute average pairwise Jensen-Shannon divergence:

\begin{equation}
    \text{Uncertainty}_{\text{JS}} = \frac{2}{K(K-1)} \sum_{i<j} D_{\text{JS}}\left(p^{(i)}(\mathbf{h}) \| p^{(j)}(\mathbf{h})\right),
\end{equation}
where $D_{\text{JS}}(p \| q) = \frac{1}{2} D_{\text{KL}}(p \| m) + \frac{1}{2} D_{\text{KL}}(q \| m)$ with $m = \frac{1}{2}(p+q)$. High divergence signals inconsistent beliefs and unreliable outputs. We adopt the Jensen-Shannon (JS) divergence due to its symmetric and bounded nature \([0, \log 2]\), which provides a robust, smoothed metric that more reliably quantifies the semantic shift in belief states induced by input perturbations or dropout noise.

To robustly aggregate features, we apply the Hodges-Lehmann estimator. While the sample mean of feature representations is susceptible to distortion by outlier hallucinations, Hodges-Lehmann estimator offers a breakdown point which ensures that the aggregated feature vector $\hat{\theta}$ remains anchored to the central tendency of the model's plausible interpretations, thereby providing a more robust than mean-based aggregation against outlier hallucinations.

\begin{equation}
    \hat{\theta} = \text{median}\left\{\frac{\mathbf{h}^{(i)} + \mathbf{h}^{(j)}}{2} \;\Big|\; 1 \leq i < j \leq K\right\}
\end{equation}

We further propose adaptive dropout regulation. The effective dropout rate $r_{\text{dropout}} = r_0 + \gamma \cdot \Delta L$ scales with estimated information loss $\Delta L = \text{Uncertainty}_{\text{JS}} + \lambda \cdot \text{Var}_{\text{HL}}(\mathbf{h})$, introducing stochasticity when uncertainty is high to avoid premature overconfidence.

Finally, we filter pseudo-labels via uncertainty thresholding. A candidate output is retained only if $\text{Uncertainty}_{\text{JS}}(t) < \tau$, where $\tau$ is calibrated on held-out validation. Flagged outputs are refined through selective re-inference until convergence. VIR achieves 15.3\% hallucination error reduction versus direct lightweight VLM usage, enabling scalable high-fidelity annotation with minimal large-model supervision.

\section{Experiments}

In this section, we systematically evaluate the effectiveness and necessity of our proposed DIQ-H benchmark and VIR framework. We perform several carefully designed experiments to validate our claims.

\subsection{Impact of Temporal Quality Degradation}

We first investigate the necessity of introducing temporally coherent degradations by comparing VLM performance under two distinct conditions: (1) sequences of images without quality degradation (clean inputs), and (2) sequences where images undergo progressive, temporally varying degradations. 

Compared to the GPT-4o-lightweight baseline, where raw lightweight VLM outputs are directly adopted as pseudo-ground truth and yield an accuracy of 72.2\% (100\% - 27.8\%) against human-verified validation labels, the VIR-refined annotations produced by GPT-4o-strict attain an accuracy of 83.3\% (100\% - 16.7\%), representing a 15.3\% relative improvement. Experimental results reveal a significant performance drop under degraded conditions. These quantitative outcomes clearly indicate the substantial impact of temporal degradation, confirming the critical importance of modeling degradation in benchmark datasets.

\subsection{Effectiveness of Uncertainty-Guided Iterative Refinement}

To validate the effectiveness of our proposed VIR framework for ground truth annotation, we conduct experiments comparing pseudo-ground truth (GT) annotations generated with and without VIR refinement. Using a set of representative sequences from DIQ-H, we measure the accuracy and robustness of the refined GT annotations against manually verified labels. Experimental results show that GT annotations generated through VIR exhibit a significant improvement in accuracy, demonstrating an average increase of 15.3$\%$ compared to annotations without VIR. These findings highlight VIR’s capability to produce reliable and scalable annotations, reducing reliance on costly human supervision.

\begin{figure*}[!t]
\centering
\includegraphics[width=\textwidth]{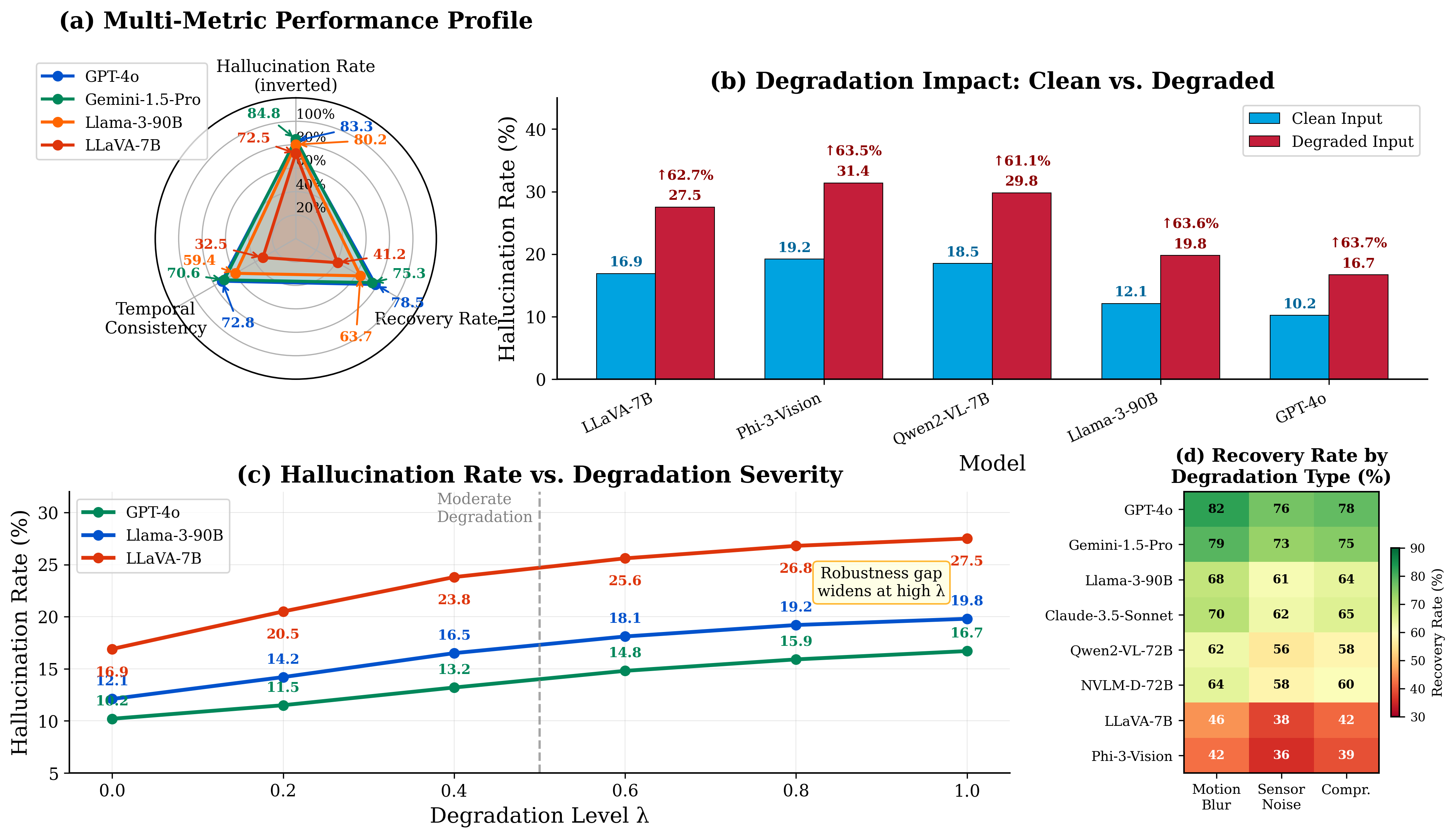}
\caption{Experimental results visualization. (a) Radar chart showing multi-dimensional performance profiles across Hallucination Rate (inverted), Recovery Rate, and Temporal Consistency for representative models. (b) Grouped bar chart comparing hallucination rates under clean vs. degraded inputs, demonstrating the substantial impact of temporal degradation. (c) Hallucination rate as a function of degradation severity $\lambda$, revealing that performance gaps between models widen under stronger corruption.}
\label{fig:results}
\end{figure*}

\begin{table*}
\centering
\caption{Evaluation of VLMs on DIQ-H Benchmark.}
\label{tab:lvlm_eval}
\resizebox{\textwidth}{!}{
\begin{tabular}{lccc}
\toprule
\textbf{Model}  & \textbf{Hallucination Rate (\%) $\downarrow$} & \textbf{Recovery Rate (\%) $\uparrow$} & \textbf{Temporal Consistency (\%) $\uparrow$} \\
\midrule
LLaVA-7B-strict  & 27.5 & 41.2 & 32.5 \\
InternVL2-8B-strict  & 25.7 & 53.6 & 47.1 \\
Phi-3-Vision-strict & 31.4 & 38.9 & 29.8 \\
Qwen2-VL-7B-strict  & 29.8 & 42.5 & 34.2 \\
Qwen2-VL-72B-strict  & 24.2 & 58.3 & 51.7 \\
Llama-3-11B-strict  & 26.5 & 55.1 & 49.3 \\
Llama-3-90B-strict  & 19.8 & 63.7 & 59.4 \\
Molmo-7B-strict  & 27.6 & 45.1 & 36.8 \\
Molmo-72B-strict  & 23.9 & 56.2 & 48.5 \\
Pixtral-12B-strict  & 26.2 & 49.8 & 42.7 \\
NVLM-D-72B-strict & 22.4 & 60.5 & 54.2 \\
\midrule
\textit{Gemini-1.5-Flash-strict}  & 18.5 & 68.9 & 63.1 \\
\textit{Gemini-1.5-Pro-strict} & \textbf{15.2} & 75.3 & 70.6 \\
\textit{Claude-3.5-Sonnet-strict}  & 19.1 & 65.2 & 60.3 \\
\textit{GPT-4o-mini-strict}  & 22.3 & 52.4 & 46.9 \\
\textit{GPT-4o-lightweight}  & 27.8 & 54.0 & 55.6 \\
\textit{GPT-4o-strict}  & 16.7 & \textbf{78.5} & \textbf{72.8} \\
\bottomrule
\end{tabular}}
\end{table*}

\subsection{Comprehensive Benchmarking of Current VLMs}

We employ three complementary metrics. Hallucination Rate quantifies unfounded assertions. Recovery Rate measures corrective capacity after errors—critical for ethical recovery in safety-critical sequences. Temporal Consistency verifies value-grounded reasoning stability across time. These metrics expose weaknesses in factual grounding, adaptive learning, and consistent ethical operation.

\begin{gather}
    \mathcal{H} = \left( 1 - \frac{N_{valid}}{N_{total}} \right) \times 100\%, \\
    Acc = (1 - \mathcal{H}) \times 100\% \\
    \mathcal{R} = \frac{1}{E} \sum_{i=1}^E \mathbb{I}(y_i^{final} = y_i^{gt}) \times 100\%, \\
    \mathcal{T} = \frac{1}{T} \sum_{j=1}^T \mathbb{I}(o_j \models \phi_j) \times 100\%,
\end{gather}

As shown in Table~\ref{tab:lvlm_eval}, GPT-4o and Gemini-1.5-Pro demonstrate superior robustness in error recovery and temporal consistency, indicating stronger value stability under dynamic degradation. LLaVA-7B and Phi-3-Vision exhibit high hallucination rates and poor recovery, revealing vulnerability to persistent error propagation in continuous sequences. DIQ-H provides a critical tool for evaluating safety-relevant capabilities in real-world deployment scenarios.

\section{Conclusion}

In this study, we presented a rigorous evaluation of modern VLMs through the DIQ-H benchmark, introducing three specialized metrics to assess critical capabilities in factual grounding, error recovery, and temporal reasoning. Our experiments demonstrated significant performance variations across both open-source and proprietary models, with GPT-4o showing particular strength in recovery capabilities (78.5$\%$) and temporal consistency (72.8$\%$), while Gemini-1.5-Pro achieved the lowest hallucination rate (15.2$\%$). The results reveal that current models still struggle with complex spatiotemporal reasoning and consistent self-correction, particularly in open-source implementations where the best temporal consistency score reached only 59.4$\%$.

These findings highlight two fundamental challenges for future research: (1) the need for more robust internal consistency mechanisms to reduce hallucination without sacrificing creative reasoning, and (2) the development of architectures capable of maintaining long-term temporal dependencies in multi-event scenarios. The demonstrated correlation between model scale and recovery performance suggests promising directions for parameter-efficient adaptation techniques. Our metric framework provides a foundation for these advancements by quantitatively isolating specific failure modes that remain obscured in conventional accuracy measurements. Future work will extend this evaluation to dynamic multi-modal interactions and real-time learning scenarios.

\bibliographystyle{splncs04}
\bibliography{references}
\end{document}